\documentclass{article}

\usepackage[preprint]{neurips_temp}

\usepackage[utf8]{inputenc} % allow utf-8 input
\usepackage[T1]{fontenc}    % use 8-bit T1 fonts
\usepackage{hyperref}       % hyperlinks
\usepackage{url}            % simple URL typesetting
\usepackage{booktabs}       % professional-quality tables
\usepackage{amsfonts}       % blackboard math symbols
\usepackage{amsmath}       % blackboard math symbols
\usepackage{nicefrac}       % compact symbols for 1/2, etc.
\usepackage{microtype}      % microtypography
\usepackage{xcolor}         % colors
\usepackage{graphicx}
\usepackage{multirow}
\usepackage{float}
\usepackage{caption}
\usepackage{adjustbox}
\usepackage{todonotes}
\usepackage{enumitem}

\title{Falcon Mamba: The First Competitive Attention-free 7B Language Model}

\author{
    Jingwei Zuo\textsuperscript{*} \And Maksim Velikanov \And Dhia Eddine Rhaiem \And Ilyas Chahed \AND Younes Belkada \And Guillaume Kunsch \And Hakim Hacid \And \\
    Technology Innovation Institute, Abu Dhabi, United Arab Emirates \\  \\
    \textsuperscript{*}\texttt{Falcon-LLM[at]tii[dot]ae}
}

\begin{document}

\maketitle

\begin{abstract}

In this technical report, we present \texttt{Falcon Mamba 7B}, a new base large language model based on the novel Mamba architecture. \texttt{Falcon Mamba 7B} is trained on 5.8 trillion tokens with carefully selected data mixtures. As a pure Mamba-based model, \texttt{Falcon Mamba 7B} surpasses leading open-weight models based on Transformers, such as Mistral 7B, Llama3.1 8B, and Falcon2 11B. It is on par with Gemma 7B and outperforms models with different architecture designs, such as RecurrentGemma 9B, and RWKV-v6 Finch 7B/14B. Currently, \texttt{Falcon Mamba 7B} is the best-performing Mamba model in the literature at this scale, surpassing both existing Mamba and hybrid Mamba-Transformer models, according to Open LLM Leaderboard \citep{open-llm-leaderboard-v2}.
Due to its architecture, \texttt{Falcon Mamba 7B} is significantly faster at inference and requires substantially less memory for long sequence generation. Despite recent studies suggesting that hybrid Mamba-Transformer models outperform pure architecture designs, we demonstrate that even the pure Mamba design can achieve similar, even superior results compared to the Transformer and hybrid designs. We make the weights of our implementation of \texttt{Falcon Mamba 7B} publicly available on \url{https://huggingface.co/tiiuae/falcon-mamba-7b}, under a permissive license\footnote{\url{https://falconllm.tii.ae/falcon-mamba-7b-terms-and-conditions.html}}.

\end{abstract}

\section{Introduction}
%Motivation
Modern foundation models are predominantly based on the Transformer and its core attention layer ~\citep{vaswani2017attention}. Due to its quadratic complexity regarding the window length, recent research attempts to propose more efficient alternatives of vanilla attention, such as FlashAttention~\citep{dao2022flashattention,dao2023flashattention2}, sliding window attention~\citep{beltagy2020longformer}. New architectures beyond Transformers such as Griffin~\citep{de2024griffin}, RWKV~\citep{peng2023rwkv}, and Mamba~\citep{gu2023mamba} have recently been proposed and have demonstrated performance comparable to Transformers. However, most of them either proved their performance at small scale, or still show a performance gap with recent Transformer-based performing LLMs.

%Related work - Mamba and Hybrid Mamba 
There have been efforts from the community to scale up Mamba LLMs beyond the original test-purpose 2.8B Mamba model~\citep{gu2023mamba}. Notable examples include Mamba-7B-rw~\citep{Mercat2024Linearizing}, Zamba 7B~\citep{glorioso2024zamba}, Samba 3.8B~\citep{ren2024samba}, Mamba2 8B (hybrid/non-hybrid)~\citep{waleffe2024empirical}. Most of these models adopt a hybrid Mamba-Transformer design, demonstrating superior performance compared to pure Transformer models. However, it remains unclear whether a pure attention-free model can match the performance of highly optimized Transformers at large data and model size scales.
%However, it remains unclear whether a pure mamba model at a large scale can achieve or surpass the performance of Transformers. 

%Brief presentation of Falcon Mamba 7B
We introduce Falcon Mamba 7B \;---\; a base (pre-trained) model with pure mamba architecture design, and the first State Space Language Model (SSLM) in the FalconLLM series. We argue that Falcon Mamba 7B answers the above question positively, and, to the best of our knowledge, the first model to do so. As measured by Open LLM Leaderboard \citep{open-llm-leaderboard-v2} collection of benchmarks, Falcon Mamba 7B matches or surpass powerful transformer-based pretrained LLMs such as Llama3.1 8B~\citep{dubey2024llama}, Mistral 7B~\citep{jiang2023mistral} and Falcon2 11B~\citep{malartic2024falcon2}.
%We built Falcon Mamba 7B, based on the pure Mamba architecture. It is the first State Space Language Model (SSLM) in the FalconLLM series, and is considered as the first competitive 7B attention-free model in the world. Falcon Mamba 7B demonstrates that even a pure Mamba design can match or surpass powerful transformer-based LLMs such as Llama3.1 8B~\citep{dubey2024llama}, Mistral 7B~\citep{jiang2023mistral} and Falcon2 11B~\citep{malartic2024falcon2}. 
Moreover, it outperforms models with other architectural designs, such as RecurrentGemma 9B~\citep{botev2024recurrentgemma} based on Griffin and RWKV-v6 Finch 7B and 14B~\citep{peng2024eagle}. 
More importantly, with the pure Mamba design, Falcon Mamba 7B maintains constant memory cost regardless of the context length, while providing extreme efficient inference for extreme long context data generation.

In this technical report, we provide a detailed overview of the model architecture, training recipes and pretraining data preparations for Falcon Mamba 7B. This will be followed by detailed comparisons with LLMs with different architecture designs on popular benchmarks. Finally, we show the broader implications of Falcon Mamba 7B, its limitations and advantages, and conclusions.

\section{Model Architecture}
The Falcon Mamba 7B model architecture is based on Mamba~\citep{gu2023mamba}. The core parameters of the architectures are summarized in Table~\ref{tab:model_params}.

\begin{table}[!htbp]
\centering
\caption{Model Parameters of Falcon Mamba 7B}
\label{tab:model_params}
\begin{adjustbox}{max width=\textwidth}
\begin{tabular}{cccccccccc}
\toprule
Params & \text{n\_layers} & \text{d\_model} & \text{exp. factor E} &\text{vocab\_size} &\text{tied\_embedding} &\text{d\_conv}  & \text{$\Delta$ proj.} size  & \text{state dim. (N)} \\ %& \text{seq\_len}\\
\midrule
7.27B &   64     & 4096     & 2     & 65024    & False   & 4   & 16   & 16  \\ %& 2048-8096\\
\bottomrule
\end{tabular}
\end{adjustbox}
\end{table}

We have untied the input embeddings from the output weights throughout the entire training process to increase model flexibility. Based on our experimental results, this approach has led to improved model performance at the 7B scale. 

Note that, in contrast to transformers, the sequence length is not a part of Mamba architecture. Any sequence length can be used during inference, while the actual ability of the model to process long sequences is determined by the sequence length used for training.

\textbf{Design decision:} Recent work~\citep{dao2024transformers,lieber2024jamba} suggests that a hybrid architecture, with interleaved attention and SSM layers, can outperform pure Transformer or SSM models. This improvement is hypothesized to arise from the complementary features from both models: the general sequence-to-sequence mapping capabilities of SSMs and the fast retrieval properties of attention layers.
%In particular, handling the limited in-context learning capability of SSM models. 
Recent Mamba-based language models follows this intuition and scale up the hybrid design beyond 2.8B models, such as Samba 3.8B~\citep{ren2024samba}, Zamba 7B~\citep{glorioso2024zamba}, Jamba 12B/52B~\citep{lieber2024jamba}. However, introducing attention layers compromises the linear scalability of the Mamba architecture, prompting the question: can a purely Mamba-based design achieve competitive performance against state-of-the-art (SoTA) open LLMs at scale, while conserving its linear scalability? 
Recent attention-free models, such as RWKV-v6~\citep{peng2024eagle}, show their performance at small scale or/and on certain academic benchmarks. However, they are far behind popular LLMs when setting up more thorough comparisons on various benchmarks.
%As in-context learning ability will emerge when model scales up, thus current 3B mamba models can not provide clear insights on this matter.

\textbf{Model stability} During pre-training, we observed consistent loss spikes that occurred randomly and unpredictably. Notably, when we applied higher learning rates, the model exhibited more pronounced loss spikes and became more prone to divergence. This phenomenon was also observed in the training of Falcon2~\citep{malartic2024falcon2}, and recent papers like Jamba~\citep{lieber2024jamba} and Mamba2~\citep{dao2024transformers} have reported similar issues. In particular, we found that the Mamba architecture is more sensitive to learning rates than Transformers. Careful model initializations and reducing model's learning rate sensibility are crucial for addressing this issue. Aligned with ~\citep{dehghani2023scaling}, it's becoming a common practice to apply pre-norm and post-norm with RMSNorm in each Transformer block to stabilize the pre-training~\citep{team2024gemma1,yang2024qwen2}. Similarly, we add RMSNorm layers after B, C and $\Delta$. From our experiments, it appears to bring a more stable training loss than other settings, such as putting an RMSNorm layer in each block before the final output projection~\citep{dao2024transformers}. This is aligned with the Jamba model designs~\citep{lieber2024jamba}.
\section{Pre-training}

\subsection{Training stategy}
Falcon-Mamba-7B was trained on 256 H100 80GB GPUs for the majority of the training, using only Data Parallelism (DP=256). This was combined with ZeRO optimization to efficiently manage memory and training processes. 

The model was trained using the AdamW optimizer with $\beta_1=0.9$ and $\beta_2=0.95$, $\epsilon=10^{-8}$, and weight decay value $0.1$. Although we didn't apply Z-loss on output logits during Falcon-Mamba-7B pre-training, in the follow-up experiments we observed that it helps to stabilize the training, in agreement with \citep{wortsman2024smallscale}. 

We applied warmup-stable-decay (WSD) learning rate schedule~\citep{hu2024minicpm} with a fixed warmup duration of 1GT, and learning rate $\eta_{\text{max}} = 6.4 \times 10^{-4}$ during the stable stage. This way, our model was trained with a relatively high learning rate during most of the pertaining, leading to a quick adaptation to data distribution shifts introduced between different training stages and the beginning of the decay stage (see section~\ref{sec:data_mix}). In the decay stage, we reduced learning rate to the minimal value $\eta_{\text{min}} = \frac{\eta_{\text{max}}}{256}$ using exponential schedule with profile $\eta(t)=\eta_\mathrm{max} \operatorname{exp}\big[-\frac{t}{t_\mathrm{decay}}\log \frac{\eta_\mathrm{max}}{\eta_\mathrm{min}}\big]$, where $t_\mathrm{decay}$ is the duration of the decay stage. Contrary to most technical reports, we found out that longer LR decay stage provided better results evaluation-wise. We kept around $10\%$ of the total training tokens for the decay to have optimal performances, which is aligned with recent miniCPM's conclusions~\citep{hu2024minicpm}.

In the beginning of the training, we used batch size rampup. Specifically, we were linearly increasing the batch size initial value $b_{\text{min}} = 128$ to the maximum value $b_{\text{max}} = 2048$ over the first 50GT. In our experiments, we noticed that batch size rampup affects the loss curve and final model performance. This effect is most conveniently interpreted in terms of gradients \emph{noise temperature} $T_{\text{noise}}$, defined for Adam optimizer as  \citep{malladi2022on}
\begin{equation}\label{eq:noise_temp}
    T_{\text{noise}} = \frac{\eta}{\sqrt{b}}.
\end{equation}
During batch size rampup, noise temperature \eqref{eq:noise_temp} is decreased. This leads to better loss during the stable LR phase but a smaller loss boost within LR decay phase. To counter this deficiency, we apply \textit{batch scaling}: keeping the Adam noise $\frac{\eta}{\sqrt{b}}$ temperature constant by adjusting learning rate $\eta$ whenever batch size $b$ is changed. We have found that batch scaling leads to a better final loss after the LR decay stage, even during long training durations much exceeding the length of rampup period. 

\subsection{Pre-training data}

%Data sources
Falcon Mamba 7B was mostly trained on the data from Falcon2-11B~\citep{malartic2024falcon2}. Since a 7B model may not be sufficient to perform promising performances on multilingual tasks without harming the English ones, we exclude multilingual data from the pre-training corpus. Nevertheless, a continual pre-training stage can be adopted to empower the model with multilingual capabilities. We adopt the same tokenizer as the \textit{Falcon} series model~\citep{almazrouei2023falcon} with no change. 

\subsubsection{Data sources}
The model was trained on a diverse data mixture consisting primarily of web, curated, code, and math data.

\textbf{Web data} We mainly leveraged \texttt{RefinedWeb}~\citep{penedo2023refinedweb}, which is a high-quality English pre-training dataset composed of five trillion tokens coming from web data only. Starting from raw Common Crawl data, samples were filtered out through language identification, filtering (line-wise and document-wise) as well as fuzzy and exact deduplication.

\textbf{Curated data} The curated dataset includes books, scientific publications (e.g., arXiv, PubMed), patents (USPTO), and conversations from platforms like Reddit, StackExchange, and Hackernews. To properly handle conversation trees, we applied the same method as in \citep{malartic2024falcon2} to enforce causal temporality, ensuring that each conversation was used only once during training.

\textbf{Code} Samples were taken from \texttt{The Stack} \citep{kocetkov2022stack} and passed through the same processing pipelines as used for web data. Code data were gradually injected during pretraining, along with continuous data collection and processing. %For pre-training, only a subset of 11 of the most popular languages was used, while for fine-tuning this number was raised to 43 to get more coding capability diversity. 

\textbf{Math} We used \texttt{Proof-Pile-2} \citep{azerbayev2023llemma} without further refinement, along with math data filtered from web using a \textit{FastText} classifier trained on \texttt{Proof-Pile-2}. 

%Removed as it's more for engineering details
%No padding tokens was used during pre-training to maximize useful compute on hardware processors. If the size of a sample is higher than context length then the sample is split, otherwise is is aggregated with others samples inside a single context length where all samples are separated by an EOS token. Apart from this EOS token, no explicit mechanism to prevent sample from attending others samples inside a sequence length was used. %this may have an impact for Mamba models 

\subsubsection{Data mixtures}\label{sec:data_mix}

\begin{figure}
    \centering
    \includegraphics[width=1\linewidth]{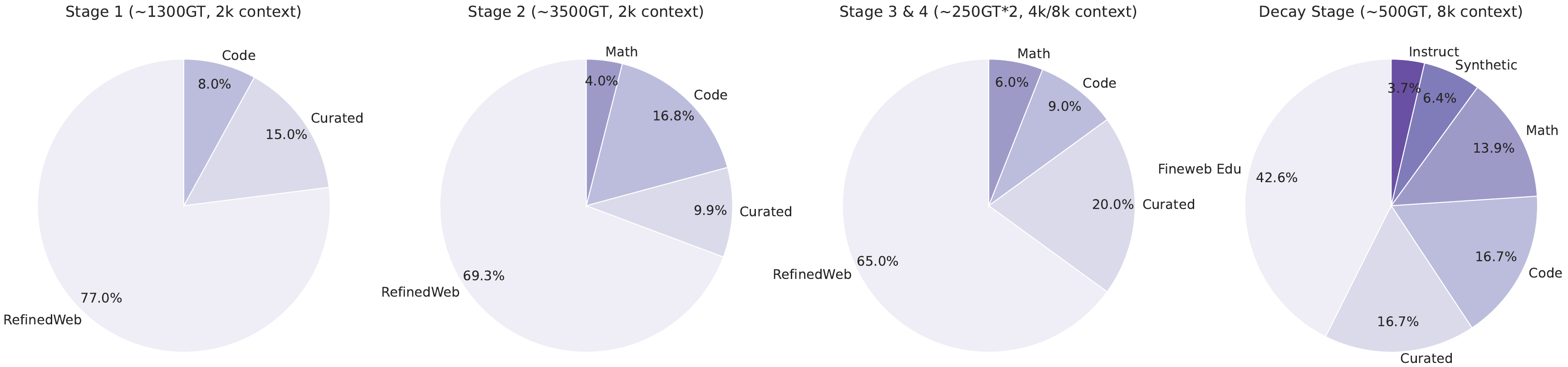}
    \caption{Data mixtures across training stages}
    \label{fig:data_mixture}
\end{figure}
% \begin{figure}
%     \centering
%     \includegraphics[width=1\linewidth]{plots/data_mixture_7b.pdf}
%     \caption{Data mixtures across training stages}
%     \label{fig:data_mixture}
% \end{figure}
%Various stages
The pre-training was conducted in four constant learning rate (LR) stages, followed by a final LR decay stage. The first four stages consisted in progressively increasing the sequence length, from 2048 up to 8192. Following the curriculum learning concept, we carefully selected data mixtures throughout the training stages as shown in Fig.~\ref{fig:data_mixture}, considering both data diversity and complexity. The main idea is to increase high quality and scientific data at late stages. Due to limited data from certain resources, we applied multiple epochs for less-represented data, e.g., math, code, curated data. 
Since the packing tokens were used in the pretraining, we carefully selected the proportions of short and long samples at each stage to prevent any distribution shifts.

%In the last stage, proportions of sample with more than 2048 tokens is 60\% in the mixture.
In the decay stage, we introduced more diverse and higher-quality data to refine or shapen the knowledge learned during earlier stages. This included using parts of Fineweb-Edu~\citep{penedo2024fineweb} as web data, along with synthetic data from Cosmopedia~\citep{benallal2024cosmopedia}. Additionally, a small portion of multitask instruction data (four epochs for 3.7\%) was used, similar to other studies~\citep{hu2024minicpm,yang2024qwen2}, to enhance the model’s zero-shot and few-shot learning capabilities.
The inclusion of instruction data during pretraining is a debated topic, as it may potentially reduce a model's fine-tuning flexibility. However, from our experimental results, we found that keeping a minimal amount of instruction data enhances Mamba's in-context retrieval ability~\citep{wen2024rnns} while not overfitting the multitask data with limited epochs of repetitions. Additionally, we observed that the training loss was still decreasing at the end of Stage 4, suggesting that the model's performance could be further improved with continued training on more high-quality data. To support the community in further research or continual training on the model, we decided to release as well the pre-decay checkpoint~\footnote{\url{https://huggingface.co/tiiuae/falcon-mamba-7b-pre-decay}} of the model.

% \section{Fine-tuning}

% Pretrained large language models frequently exhibit misalignment with human behavior, limiting their effectiveness as AI assistants in numerous applications. Nevertheless, recent research indicates that alignment strategies, such as supervised fine-tuning (SFT) and reinforcement learning from human feedback (RLHF), can enhance their utility.

% Consequently, in addition to the base model, \textit{Falcon Mamba 7B}, we introduce an instruct variant, \textit{Falcon Mamba 7B Instruct}.

% \subsection{Training}

% To simplify the experimental setup and demonstrate that the Mamba architecture retains the adaptive capabilities of Transformer-based models, we utilize SFT as the sole fine-tuning method. 

% \textit{Falcon Mamba 7B Instruct} was fine-tuned with only a marginal increase in gigatokens compared to \textit{Falcon Mamba 7B}. These additional tokens were sourced exclusively from instruct datasets consisting of user prompts and corresponding answers. Although some of these tokens may have already been encountered by the model during the LR decay phase, a specialized chat template was employed at this stage to emulate a conversational context. Note that \textit{Falcon Mamba 7B Instruct} chat template doesn't use a system prompt which makes it easily tunable by the community.

% In our training protocol, we apply loss masking to both system and user inputs, ensuring that the model is fine-tuned based solely on the provided responses.

\section{Evaluation and Results}

\subsection{Benchmark results}

We conducted a comparative evaluation of our model against state-of-the-art models across three distinct architectural categories: State Space Models (SSMs), Transformers, and Hybrid models. The Hybrid models integrate a combination of attention mechanisms with Recurrent/Mamba blocks.

Benchmarks were selected where results are publicly available and independently conducted by HuggingFace, which span a broad range of top-level categories to assess the model’s versatility and performance across various tasks:
\begin{itemize}%[leftmargin=*]
    \item Instruction following: IFEval (0-shot)~\citep{zhou2023instruction}
    \item Math, reasoning, and problem-solving: GSM8K (5-shots)~\citep{cobbe2021training}, MATH-Lvl5 (4-shots)~\citep{hendrycks2021measuring}, ARC Challenge (25-shots)~\citep{clark2018think}, GPQA  (0-shot)~\citep{rein2023gpqa}, MuSR (0-shot)~\citep{sprague2023musr}
    \item Aggregate: MMLU (5-shots)~\citep{hendrycks2020measuring}, MMLU-Pro (5-shots)~\citep{wang2024mmlu}, BIG-Bench Hard (BBH) (3-shots)~\citep{suzgun2022challenging}
\end{itemize}

As shown in Table~\ref{tab:perf_hf_v1} and Table~\ref{tab:perf_hf_v2}, wherever possible, we extracted results for competitor models from the HF Leaderboards v1\citep{open-llm-leaderboard-v1} and v2\citep{open-llm-leaderboard-v2}, ensuring an unbiased comparison. When leaderboard results were unavailable, we used the best available results, either from reported findings or our internal evaluations. Internal evaluations were performed using the \texttt{lm-evaluation-harness}~\citep{eval-harness} and \texttt{lighteval}~\citep{lighteval} packages.

% \begin{table}[H]
% \centering
% \begin{tabular}{|l|l|}
% \hline
% \textbf{Category} & \textbf{Benchmark(s)} \\ \hline
% \multirow{1}{*}{Commonsense reasoning/understanding} & WinoGrande (Sakaguchi et al., 2021) \\
% & HellaSwag (Zellers et al., 2019) \\
% & TruthfulQA (Lin et al., 2019) \\\hline
% \multirow{3}{*}{Math, reasoning, and problem solving} & GSM8K (Cobbe et al., 2021)\\
%                                                       & MATH (Hendrycks et al., 2021b) \\
%                                                       & ARC Challenge (Clark et al., 2018) \\
%                                                       & GPQA (Rein et al., 2023) \\ \hline
% \multirow{4}{*}{Aggregate}             & MMLU (Hendrycks et al., 2021a) \\
%                                        & MMLU-Pro (Wang et al., 2024b) \\
%                                        & BIG-Bench Hard (Suzgun et al., 2023) \\
%                                        & MuSR (Sprague et al., 2023) \\
%                                        \hline
% \multirow{1}{*}{Instruction following}  & IFEval (Zhou et al., 2023) \\
%                                        \hline
% \end{tabular}
% \caption{Benchmark categories and corresponding datasets}
% \label{tab:benchmarks}
% \end{table}

\begin{table}[ht]
\centering
\caption{Model Performance on HF Leaderboard v1 tasks: \textbf{bold} (best), \underline{underline} (second best)}
\label{tab:perf_hf_v1}
\begin{adjustbox}{max width=\textwidth}
\begin{tabular}{lcccccccc}
\toprule
\textbf{Model Name} & \textbf{ARC-25} & \textbf{HellaSwag-10} & \textbf{MMLU-5} & \textbf{Winogrande-5} & \textbf{TruthfulQA-0} & \textbf{GSM8K-5} & \textbf{Average} \\

\midrule

\multicolumn{8}{l}{\textbf{RWKV models}} \\
RWKV-v6-Finch-7B* & 43.86 & 75.19 & 41.69 & 68.27 & 42.19 & 19.64 & 48.47 \\
RWKV-v6-Finch-14B* & 47.44 & 78.86 & 52.33 & 71.27 & 45.45 & 38.06 & 55.57 \\
\midrule
\multicolumn{8}{l}{\textbf{Transformer models}} \\
Falcon2-11B & 59.73 & \underline{82.91} & 58.37 & 78.30 & \underline{52.56} & \underline{53.83} & \textbf{64.28} \\
Meta-llama-3-8B & 60.24 & 82.23 & \textbf{66.70} & 78.45 & 42.93 & 45.19 & 62.62 \\
Meta-llama-3.1-8B & 58.53 & 82.13 & \underline{66.43} & 74.35 & 44.29 & 47.92 & 62.28 \\
Mistral-7B-v0.1 & 59.98 & \textbf{83.31} & 64.16 & 78.37 & 42.15 & 37.83 & 60.97 \\
Mistral-Nemo-Base-2407 (12B) & 57.94 & 82.82 & 64.43 & 73.72 & 49.14 & \textbf{55.27} & 63.89 \\
Gemma-7B & \underline{61.09} & 82.20 & 64.56 & \underline{79.01} & 44.79 & 50.87 & 63.75 \\
\midrule
\multicolumn{8}{l}{\textbf{Hybrid SSM-attention models}} \\
RecurrentGemma-9b** & 52.00 & 80.40 & 60.50 & 73.60 & 38.60 & 42.60 & 57.95 \\
Zyphra/Zamba-7B-v1* & 56.14 & 82.23 & 58.11 & \textbf{79.87} & 52.88 & 30.78 & 60.00 \\
\midrule
\multicolumn{8}{l}{\textbf{Pure SSM models}} \\
TRI-ML/mamba-7b-rw* & 51.25 & 80.85 & 33.41 & 71.11 & 32.08 & 4.70 & 45.52 \\
FalconMamba-7B (pre-decay)* & 49.23 & 80.25 & 57.27 & 70.88 & 37.28 & 21.83 & 57.29 \\
FalconMamba-7B* & \textbf{62.03} & 80.82 & 62.11 & 73.64 & \textbf{53.42} & 52.54 & \underline{64.09} \\
\bottomrule
\end{tabular}
\end{adjustbox}
\end{table}

\vspace{-1em}

\begin{table}[ht]
\centering
\caption{Model Performance on HF Leaderboard v2: \textbf{bold} (best), \underline{underline} (second best)}
\label{tab:perf_hf_v2}
\begin{adjustbox}{max width=\textwidth}
\begin{tabular}{lcccccccc}
\toprule
\textbf{Model Name} & \textbf{IFEval-0} & \textbf{BBH-3} & \textbf{Math-Lvl5-4} & \textbf{GPQA-0} & \textbf{MuSR-0} & \textbf{MMLU-PRO-5} & \textbf{Average} \\
\midrule
\multicolumn{2}{l}{\textbf{RWKV models}} \\
RWKV-v6-Finch-7B & 27.65 & 9.04 & 1.11 & 2.81 & 2.25 & 5.85 & 8.12\\
RWKV-v6-Finch-14B & 29.81 & 12.89 & 1.13 & 5.01 & 3.16 & 11.3 & 10.55 \\
\midrule
\multicolumn{8}{l}{\textbf{Transformer models}} \\
Falcon2-11B & \underline{32.61} & 21.94 & 2.34 & 2.80 & 7.53 & 15.44 & 13.78 \\
Meta-llama-3-8B & 14.55 & 24.50 & 3.25 & \underline{7.38} & 6.24 & 24.55 & 13.41 \\
Meta-llama-3.1-8B & 12.70 & \underline{25.29} & 4.61 & 6.15 & 8.98 & \underline{24.95} & 13.78 \\
Mistral-7B-v0.1 & 23.86 & 22.02 & 2.49 & 5.59 & 10.68 & 22.36 & 14.50 \\
Mistral-Nemo-Base-2407 (12B) & 16.83 & \textbf{29.37} & \underline{4.98} & 5.82 & 6.52 & \textbf{27.46} & \underline{15.08} \\
Gemma-7B & 26.59 & 21.12 & \textbf{6.42} & 4.92 & \textbf{10.98} & 21.64 & \textbf{15.28} \\
\midrule
\multicolumn{8}{l}{\textbf{Hybrid SSM-attention models}} \\
RecurrentGemma-9b & 30.76 & 14.80 & 4.83 & 4.70 & 6.60 & 17.88 & 13.20 \\
Zyphra/Zamba-7B-v1* & 24.06 & 21.12 & 3.32 & 3.03 & 7.74 & 16.02 & 12.55 \\
\midrule
\multicolumn{8}{l}{\textbf{Pure SSM models}} \\
TRI-ML/mamba-7b-rw* & 22.46 & 6.71 & 0.45 & 1.12 & 5.51 & 1.69 & 6.25 \\
FalconMamba-7B (pre-decay)* & 24.05 & 11.01 & 1.71 & 3.05 & 8.68 & 8.59 & 9.52 \\
FalconMamba-7B & \textbf{33.36} & 19.88 & 3.63 & \textbf{8.05} & \underline{10.86} & 14.47 & 15.04 \\
\bottomrule
\end{tabular}
\end{adjustbox}
\caption*{\footnotesize \textbf{Note:} * indicates internal evaluations, ** denotes results taking from paper or model card.}
\end{table}

%overall comparison with other architectures with different model sizes
Globally, Falcon-Mamba-7B outperforms models of similar scale, regardless of architecture, including Transformer models (Llama3/3.1-8B~\citep{dubey2024llama}, Mistral-7B~\citep{jiang2023mistral}), RWKV-v6-Finch-7B~\citep{peng2024eagle}, and hybrid models like Zamba-7B, as well as Mamba-7B-RW~\citep{Mercat2024Linearizing}. Furthermore, it either outperforms or is comparable to larger models, such as Falcon2-11B~\citep{malartic2024falcon2}, RWKV-v6-Finch-14B~\citep{peng2024eagle}, Gemma-7B (8.54B)~\citep{team2024gemma1}, RecurrentGemma-9B~\citep{botev2024recurrentgemma}, and Mistral-Nemo-12B~\footnote{\url{https://huggingface.co/mistralai/Mistral-Nemo-Base-2407}}. This positions Falcon-Mamba-7B as the first competitive attention-free 7B model in the community, with promising performance across a variety of tasks. We also report the model's performance at the pre-decay checkpoint, with a notable performance boost observed during the decay stage. The decay stage can provide valuable insights for determining data mixtures in larger-scale models and simulating a condensed pretraining phase.

%unique features of mamba7b
While recent studies~\citep{waleffe2024empirical,wen2024rnns} indicate that pure Mamba/Mamba2 designs lag behind Transformers in tasks like copying and in-context learning, Falcon-Mamba-7B has shown promising performance in few-shot learning tasks, such as MMLU, ARC-25, and GSM8K. This suggests that the quality of data and training strategies during pretraining play a more crucial role than the architecture itself and can mitigate these potential disadvantages. Moreover, Falcon-Mamba-7B excels in long-context reasoning tasks, e.g., MuSR~\citep{sprague2023musr}, highlighting its significant potential in long-context learning scenarios.

\subsection{Throughput and memory consumption}
The attention mechanism is inherently limited in processing long sequences due to the increasing compute and memory costs as sequence length grows. Leveraging the theoretical efficiency of SSM models in handling large sequences~\citep{gu2023mamba}, Falcon-Mamba-7B demonstrates that these scaling limitations can indeed be overcome without compromising performance.

\textbf{Setup} To replicate real-world use cases, we compared the memory usage and generation throughput of Falcon-Mamba-7B with popular Transformer-based models of a similar scale, including Llama3.1-8B~\citep{dubey2024llama}, Mistral-7B~\citep{jiang2023mistral}, and Qwen2-7B~\citep{yang2024qwen2}. All evaluations were conducted using the Hugging Face \texttt{transformers} library~\citep{Wolf_Transformers_State-of-the-Art_Natural_2020}.
% \textbf{Setup} To replicate possible real-world use cases,  we perform a comparison of memory usage and generation throughput between Falcon-Mamba-7B and popular Transformer-based models at the same scale, including Llama3.1-8B~\citep{dubey2024llama}, Mistral-7B~\citep{jiang2023mistral}, Qwen2-7B~\citep{yang2024qwen2}. We use the optimum-benchmark~\footnote{\url{https://github.com/huggingface/optimum-benchmark}} library to perform our evaluations on top of our implementation of the model integrated in Hugging Face \texttt{transformers} library~\citep{Wolf_Transformers_State-of-the-Art_Natural_2020}.
For a fair comparison, we rescaled the vocabulary size of all transformer models to match Falcon-Mamba-7B, since it has a big impact on the memory footprint of the model.

%We fix the context length to 4096 and generate 128 tokens by varying the batch size between 1 and 32. Other model implementations use PyTorch \citep{Ansel_PyTorch_2_Faster_2024} \href{https://pytorch.org/docs/stable/generated/torch.nn.functional.scaled_dot_product_attention.html}{\texttt{torch.scaled\_dot\_product\_attention}} optimized attention operation that use Flash Attention 2 \citep{dao2022flashattention} under the hood. We load the models in \texttt{float16} precision for our benchmarks.

\textbf{Parallel Prefill and Sequential Prefill}
Before diving into the results, it is important to clarify the difference between the prompt (prefill) and generated (decode) parts of a sequence. For state space models (SSMs), the prefill process is more critical than for transformer models.
When a transformer generates the next token, it must attend to the keys and values of all previous tokens in the context, resulting in both memory and generation time scaling linearly with context length. In contrast, an SSM only stores and attends to its recurrent state, which avoids the need for additional memory or time when generating large sequences. While this demonstrates the efficiency of SSMs during the decoding phase, the prefill phase requires additional framework optimizations to fully leverage the SSM architecture.

The standard method for prefill is processing the entire prompt in parallel, maximizing GPU utilization, referred to here as \textbf{Parallel Prefill}. This is the approach used in most frameworks like \texttt{Optimum-Benchmark}~\footnote{\url{https://github.com/huggingface/optimum-benchmark}}. In this approach, the memory usage grows with prompt length due to the need to store hidden states for each token. For transformers, memory is dominated by stored key-value (KV) caches, whereas SSMs don't require KV caching. However, for SSMs, the memory required to store hidden states still scales with the prompt length, making it challenging to handle arbitrarily long sequences, similar to transformers.
An alternative method, which we referred to as \textbf{Sequential Prefill}, processes the prompt token by token (or in larger chunks for better GPU usage), similar to sequence parallelism. While this method offers little benefit for transformers, it allows SSMs to process arbitrarily long prompts, mitigating the memory scaling issue seen with parallel prefill. This requires more community supports for optimizing existing inference frameworks for SSMs. 

\begin{figure}[htbp]
    \centering
    \includegraphics[width=0.6\textwidth]{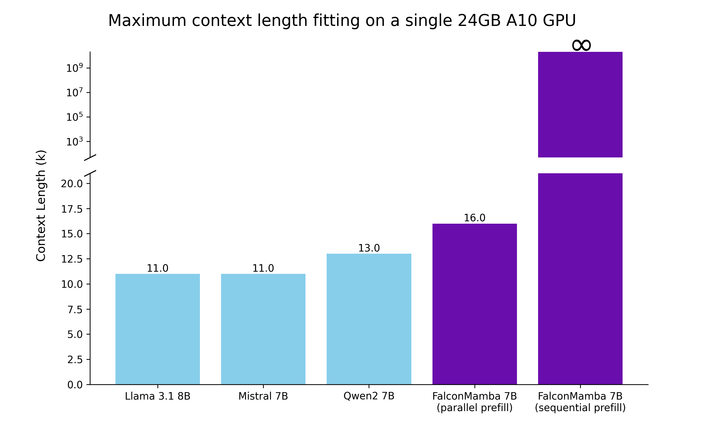}
    \caption{We vary the context length of the prompt to determine the maximum sequence length that could be processed without encountering an out-of-memory (OOM) error. To ensure a fair comparison, all models were configured with a rescaled vocabulary size.}
    \label{fig:max_length_comp}
\end{figure}

With these considerations in mind, we first evaluate the maximum sequence length that can fit on a single 24 GB A10 GPU, as shown in Fig.~\ref{fig:max_length_comp}. The batch size is fixed at 1, and we employ float32 precision for all operations. Our results show that, even for parallel prefill, Falcon-Mamba-7B is capable of fitting larger sequences compared to a standard transformer architecture, while in sequential prefill, Falcon-Mamba-7B can unlock its full potential and process arbitrarily long prompts.

\begin{figure}[htbp]
    \centering
    \includegraphics[width=1\linewidth]{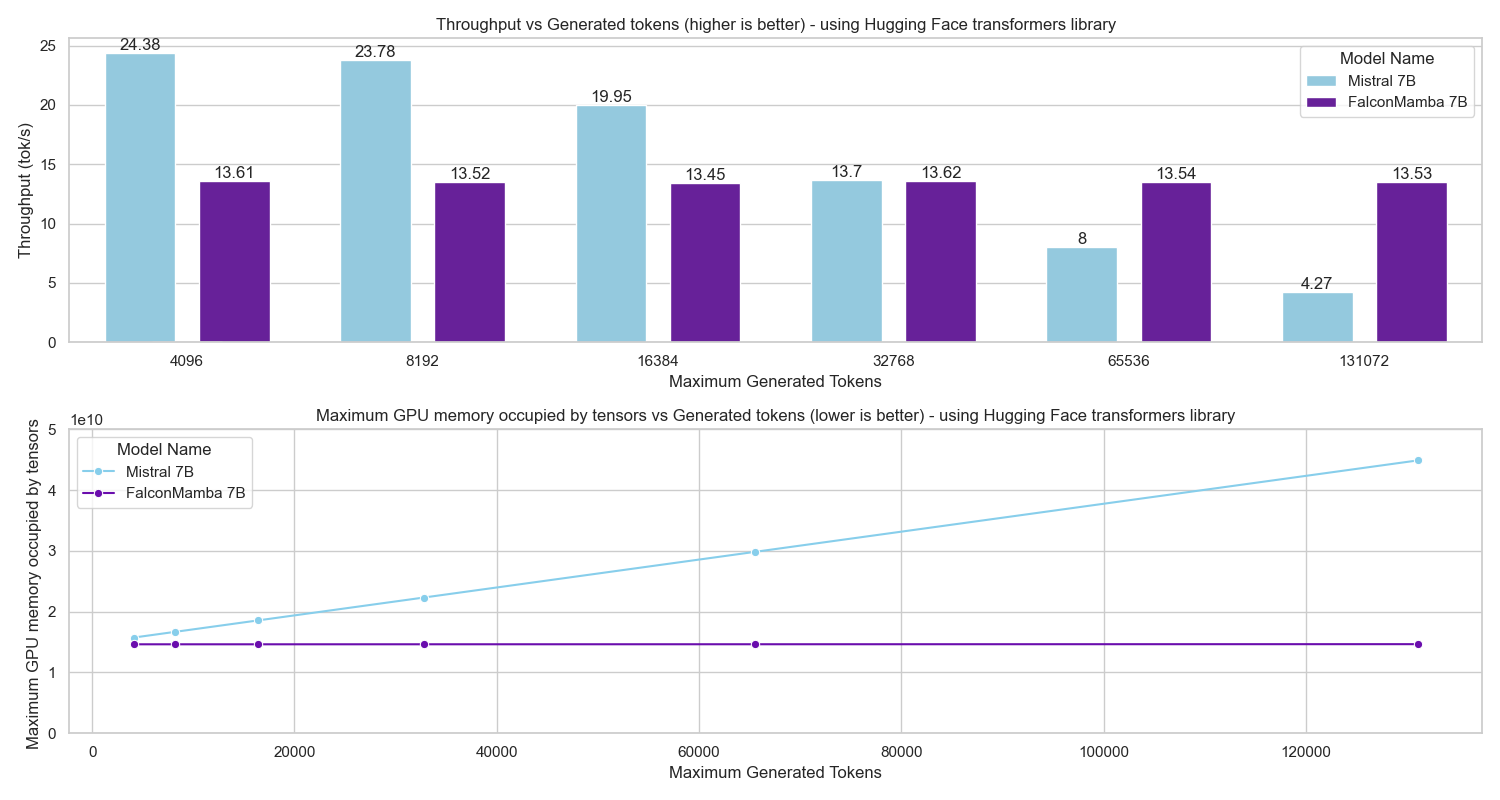}
    \caption{With a fixed batch size and context length of 1, we vary the generated tokens up to 130k for Faclon-Mamba-7B, and Mistral-7B with a resized vocabulary for fair comparisons.}
    \label{fig:thoughtput_memory}
\end{figure}

%For the record, with a batch size of 1 the maximum context length before encountering OOM failure on A10-24GB is 16,384 while it is only around 6,144 for Qwen2-7b and Llama3-8b.
Next, we evaluate the generation throughput in an extreme setting: a prompt of length 1 and up to 130k generated tokens, using a batch size of 1 on an 80GB H100 GPU. The results, reported in Fig.~\ref{fig:thoughtput_memory}, reveal that Falcon-Mamba-7B maintains a constant throughput across all generated tokens, without any increase in peak CUDA memory usage. In contrast, the Mistral-7B model exhibits a linear increase in peak memory consumption, and its generation speed decreases as the number of generated tokens grows.
%Results are reported on the figures \ref{fig:latency_4096_all}. We particularly observe that for large sequence length and batch sizes Our Mamba has a higher throughput than all existing other models with comparable number of parameters, and is generally more memory efficient compared to other models, except \texttt{mistralai/Mistral-7B-v0.1} which can be explained by the fact that the later uses sliding window attention and a vocabulary size which is smaller by a factor of two. The red cross denotes CUDA OOM failure cases.

%We confirm this hypothesis by running the same benchmarks for a random model that has the same architecture as Mistral-7B with a vocabulary size that is identical as our Mamba, and observing that our Mamba is always more memory efficient than that model that rely on sliding windows attention. Results are reported on the figure \ref{fig:latency_resized_mistral}.

%\begin{figure}[htbp]
%    \centering
%%    \includegraphics[width=0.8\textwidth]{plots/latency_throughput/plot_2048_128_all.png}
%    \caption{Our benchmark ran with a context length of 2048 and generated tokens of 128 for our Mamba, Mistral-7B, Llama-8b and Qwen2-7b model.}
%    \label{fig:latency_2048_all}
%\end{figure}

% \begin{figure}[htbp]
%     \centering
%     \includegraphics[width=0.8\textwidth]{plots/latency_throughput/plot_4096_128_all.png}
%     \caption{Our benchmark ran with a context length of 4096 and generated tokens of 128 for our Mamba, Mistral-7B, Llama-8b and Qwen2-7b model.}
%     \label{fig:latency_4096_all}
% \end{figure}

% \begin{figure}[htbp]
%    \centering
%     \includegraphics[width=0.8\textwidth]{plots/latency_throughput/plot_2048_128.png}
%     \caption{Our benchmark ran with a context length of 2048 and generated tokens of 128 for our Mamba, Mistral-7B and a Mistral-7B model with a resized vocabulary size (32000 for the default model and 65024 for the resized model)}
%     \label{fig:latency_resized_mistral}
% \end{figure}

%\begin{figure}[htbp]
%    \centering
%    \includegraphics[width=0.8\textwidth]{plots/latency_throughput/plot_1_512.png}
%    \caption{Our benchmark ran with a fixed batch size of 1 and generated tokens of 512 for our Mamba, and Qwen2-7b model. Missing bar plots denotes a CUDA out of memory error.}
%    \label{fig:latency_large_seqlen}
%\end{figure}

\section{Model Integration and Availability}

\subsection{Batched generation support}
In real-world scenarios, input sequences of varying lengths are often batched together for efficiency, which introduces padding tokens to align the sequences. This can pose challenges for SSM-based models like Mamba, as right-side padding, while effective during training—where padding tokens are masked out in the loss computation—becomes problematic during inference. In inference, the Mamba model predicts the next token based on all previous hidden states, so including padding tokens from shorter sequences can lead to inaccurate predictions.

In Transformer models, left-side padding is typically used to prevent padding tokens from interfering with the attention mechanism. For Mamba models, which use both SSMs and convolutional layers, a different approach is required. Apart from left-side padding, Falcon-Mamba-7B handles this by zeroing out the hidden states for left padding tokens both before and after the causal convolution step. This ensures padding tokens do not influence the model's predictions during generation.

\subsection{Model Availability}
%As we want the ML community to benefit from this model and be able to easily build on top of it, we made the model easily accessible within the community in order for anymore to play with the model and broadly share the built artifacts.
The Falcon-Mamba-7B models, including the pre-decay checkpoint, are made available under the Falcon Mamba 7B TII License~\footnote{\url{https://falconllm.tii.ae/falcon-mamba-7b-terms-and-conditions.html}}, a permissive Apache 2.0-based software license which includes an acceptable use policy~\footnote{\url{https://falconllm.tii.ae/falcon-mamba-7b-acceptable-use-policy.html}} that promotes the responsible use of AI. 

The models are fully integrated within the Hugging Face ecosystem and can be accessed through the Transformers library~\citep{Wolf_Transformers_State-of-the-Art_Natural_2020}. This includes support for inference, quantization (using most supported quantization schemes), and fine-tuning via the TRL library~\citep{vonwerra2022trl}. All associated artifacts, including GGUF files, can be browsed through the \href{https://huggingface.co/collections/tiiuae/falconmamba-7b-66b9a580324dd1598b0f6d4a}{Falcon Mamba 7B collection} in Hugging Face.

Additionally, support for Falcon-Mamba-7B has been added to the \texttt{llama.cpp} package~\footnote{\url{https://github.com/ggerganov/llama.cpp}}, enabling easy deployment of Falcon-Mamba-7B on local machines using CPU hardware. We are planning to expand the support for more platforms in the future.
% \begin{figure}[htbp]
%     \centering
%     \includegraphics[width=1.0\textwidth]{plots/screenshot-llamacpp.png}
%     \caption{Screenshot of generation results using \href{https://github.com/ggerganov/llama.cpp}{\texttt{llama.cpp}} quantized model}
%     \label{fig:llamacpp}
% \end{figure}

\section{Discussions and conclusion}
We have introduced Falcon Mamba 7B, the first competitive 7B language model based purely on the Mamba architecture. Our results show that it matches or outperforms state-of-the-art transformer models such as Llama 3.1 and Mistral 7B in a variety of benchmarks. This way, Falcon Mamba 7B sets a new benchmark for attention-free models, proving that pure SSM-based designs can achieve state-of-the-art performance. We hope that our model will strengthen the belief in further innovation of efficient language model architectures, challenging the infamous ``attention is all you need'' saying.

The main advantage of mamba architecture lies in the long-context generation, where it maintains constant memory and throughput usage regardless of sequence length. We have confirmed this statement with throughput and memory analysis for Falcon Mamba 7B. However, as we focused on obtaining a strong general-purpose language model, the actual proficiency of the model in long sequence understanding and generation was not emphasized in Falcon Mamba 7B training strategy, featuring rather medium 8k context length. Tailoring the training procedure towards extra-large contexts and verifying mamba proficiency in this regime remains an important yet underexplored area for future research and development. If successful, it would make mamba-based models ideal for real-world applications requiring low-latency, large-scale generation, e.g., audio, video.

While Falcon Mamba 7B performs well, particularly in reasoning tasks and long-context learning, it shows potentially some limitations in in-context learning compared to Transformers.
Although high-quality data, especially Chain-of-Thought (CoT) instruction data or tailored prompting techniques~\citep{arora2024just}, help mitigate these potential disadvantages, it may still not be sufficient to close the gap with Transformers~\citep{wen2024rnns}, given the same data budget. However, data scaling and model scaling in the Mamba architecture have been less explored in the literature, leaving the potential limitations and optimizations of Mamba as an open area for further research. Moreover, %the complementary features between Mamba and Transformers, along with their combined training dynamics and strategies, remain underexplored. These are crucial areas for developing more optimized hybrid models, 
the complementary features of sequence mixing performed by SSM and attention suggest that hybrid models might have the best of both worlds. Although many recent models~\citep{lieber2024jamba,ren2024samba,dao2024transformers,de2024griffin} have started to explore this direction, we believe that the question of how to optimally use SSM and attention in a single architecture remains open.

%Hybrid, scaling-up, training dynamics, 
%Optimizing inference frameworks for SSMs will be crucial to unlocking the full potential of this architecture, which will require collective support from the community.

\section*{Acknowledgments}
We would like to thank the Hugging Face team for their continuous support and model integration within their ecosystem. We also extend our gratitude to Tri Dao and Albert Gu for implementing and open-sourcing the Mamba architecture for the community.

%%%%%%%%%%%%%%%%%%%%%%%%%%%%%%%%%%%%%%%%%%%%%%%%%%%%%%%%%%%%
\bibliographystyle{iclr2025_conference}
\bibliography{references}

\end{document}